\documentclass[10pt,a4paper,logo]{paper}

\usepackage[all]{hypcap}

\usepackage[authoryear, round]{natbib}

\usepackage{nicefrac}       
\usepackage{multirow}
\usepackage{cleveref}
\usepackage{dsfont}
\usepackage{makecell}
\usepackage{subcaption}
\usepackage{bbm}
\usepackage{wrapfig}
\usepackage{mathtools,xparse}
\usepackage{array}
\usepackage{arydshln}
\usepackage{scalerel}
\usepackage{tablefootnote}
\usepackage{tocloft}
\usepackage{adjustbox}
\usepackage{setspace}
\usepackage[most,skins,theorems]{tcolorbox}

\newtcolorbox{promptbox}{
  colback=gray!5,
  colframe=gray!70,
  boxrule=0.8pt,
  arc=2mm,
  left=10pt,
  right=10pt,
  top=10pt,
  bottom=10pt,
  width=\linewidth,
  fontupper=\ttfamily\footnotesize,
  enhanced,
}

\newboolean{showsection}
\setboolean{showsection}{true}
\makeatletter
\@namedef{ver@everyshi.sty}{}
\makeatother

\definecolor{custom_green}{rgb}{0.0, 0.5, 0.0}
\definecolor{custom_red}{rgb}{1.0, 0.01, 0.24}
\definecolor{custom_blue}{HTML}{C9DAF7}
\definecolor{custom_purple}{HTML}{D9D1E9}
\definecolor{title_blue}{HTML}{204899}
\definecolor{cite_blue}{HTML}{044dc1}
\definecolor{cite_purple}{HTML}{7406a7}

\hypersetup{
    colorlinks = true,
    citecolor = {cite_blue},
    linkcolor = {cite_purple},
    urlcolor = {cite_purple},
}

\usepackage{algorithm}
\usepackage{algpseudocode}
\usepackage{xspace}

\definecolor{oursgray}{gray}{0.90}

\newcommand{\model}{\hat{p}}

\definecolor{blanchedalmond}{rgb}{1.0, 0.92, 0.8}
\definecolor{carmine}{rgb}{0.59, 0.0, 0.09}
\definecolor{lightblue}{rgb}{0.22,0.45,0.70}%

\renewcommand{\mathbf}{\boldsymbol}

\makeatletter
\def\Ddots{\mathinner{\mkern1mu\raise\p@
\vbox{\kern7\p@\hbox{.}}\mkern2mu
\raise4\p@\hbox{.}\mkern2mu\raise7\p@\hbox{.}\mkern1mu}}
\makeatother

\numberwithin{equation}{section}

\definecolor{amaranth}{rgb}{0.9, 0.17, 0.31}
\definecolor{antiquebrass}{rgb}{0.8, 0.58, 0.46}
\definecolor{antiquefuchsia}{rgb}{0.57, 0.36, 0.51}
\definecolor{chromeyellow}{rgb}{0.31, 0.47, 0.26}

\newcommand{\1}{\mathds 1}

\renewcommand{\model}{XFlow\xspace}

\usepackage{amsmath,amsfonts,bm}

\def\eqref#1{equation~\ref{#1}}

\def\1{\bm{1}}

\DeclareMathAlphabet{\mathsfit}{\encodingdefault}{\sfdefault}{m}{sl}
\SetMathAlphabet{\mathsfit}{bold}{\encodingdefault}{\sfdefault}{bx}{n}

\makeatletter
\def\mathcolor#1#{\@mathcolor{#1}}
\def\@mathcolor#1#2#3{%
  \protect\leavevmode
  \begingroup
    \color#1{#2}#3%
  \endgroup
}
\makeatother

\Crefformat{equation}{#2Eq.\;(#1)#3}
\Crefformat{figure}{#2Figure #1#3}
\Crefformat{assumption}{#2Assumption #1#3}
\Crefname{assumption}{Assumption}{Assumptions}

\usepackage{crossreftools}
\makeatletter
\renewcommand\footnoterule{%
  \kern 15\p@
  \hrule \@width 2in \kern 2.6\p@
  \vspace{4pt}
}
\makeatother

\title{\model: A Workflow Model for Instruction-Guided Lesion
Segmentation in Chest X-rays}

\reportnumber{}

\author[1]{Geon Choi}
\author[1]{Hangyul Yoon}
\author[2]{Hyunki Park}
\author[3]{Sang Hoon Seo}
\author[1]{Edward Choi}

\affil[1]{KAIST}
\affil[2]{Korea University Guro Hospital}
\affil[3]{Samsung Medical Center}

\correspondingauthor{Correspondence to: \email{\{choigeon, edwardchoi\}@kaist.ac.kr}.}

\begin{abstract}

Existing text-guided segmentation models in the medical domain cover only a narrow set of anatomical structures and lesions in chest X-rays (CXRs), and most of them assume that the queried target is always present in the image.
Instruction-guided lesion segmentation (ILS) was introduced to overcome these limitations by segmenting diverse lesion types from simple user instructions while also recognizing when the queried lesion is absent, and ROSALIA was proposed as the first model for this task.
However, the masks produced by ROSALIA remain of limited quality, often carrying scattered noise.
Moreover, ROSALIA predicts the mask in a single shot, which differs fundamentally from how radiologists perceive and delineate lesions in practice.
A radiologist first surveys the entire thorax, then localizes the approximate region of abnormality, and only then refines the lesion contour.
Motivated by this coarse-to-fine, multi-level perception process, we present \model, a workflow model for ILS that combines box-based localization with multi-turn point refinement.
\model detects the lungs, decides whether the queried finding is present in each of them, and prompts a fine-tuned SAM with the lesion box for an initial mask.
It then corrects that mask through point prompts until its boundary follows the lesion, leaving every intermediate decision visible.
Our experiments show that \model achieves the best segmentation quality on both internal and external evaluation.
Notably, it surpasses ROSALIA in segmentation quality even when the two are trained on the same lesion annotations. Code and model weights will be made publicly available.
\end{abstract}

\begin{document}

\maketitle

\section{Introduction}

\vspace{-0.5em}

Medical imaging plays a crucial role in modern medicine as a primary source of
visual evidence. Among the available modalities, chest X-ray (CXR) is the most
widely used examination for rapidly assessing a patient's overall condition~\citep{broder2011imaging}.
Compared with other modalities such as CT or MRI, however, CXR offers limited
spatial resolution and projects three-dimensional anatomy onto a single plane,
so that multiple structures overlap within the same region. Accurate
interpretation therefore demands substantial radiological expertise. In routine
clinical practice, however, radiologists must read a large volume of studies
each day and are often expected to interpret an individual case within minutes.
Among the required reading tasks, lesion segmentation requires pixel-level
judgment rather than a mere image-level impression, and this step consumes the
largest share of a radiologist's reading time.

A text-guided segmentation model could ease this burden, yet existing medical models cover only a limited range of anatomical structures and lesion types in CXRs~\citep{huang2024cross, li2023lvit}, since collecting large-scale, high-quality annotations from radiologists across diverse findings is time-intensive.
Moreover, most assume the queried target is present in the image, whereas radiologists must first determine whether a lesion exists at all.
To address these limitations, \citet{choi2026instruction} introduced instruction-guided lesion segmentation (ILS), which segments diverse CXR lesions from simple instructions while also recognizing when the queried lesion is absent.
Along with the task, they released MIMIC-ILS, a large-scale lesion segmentation dataset~\citep{PhysioNet-mimic-cxr-ext-ils-1.0.0}, and proposed ROSALIA, the first vision-language model (VLM) for ILS.
That work, however, centers on the dataset, and ROSALIA's masks remain
of limited quality, often carrying scattered noise.

We attribute these failures to ROSALIA's architecture: it adopts LISA~\citep{lai2024lisa}, which couples a VLM with the Segment Anything Model (SAM)~\citep{kirillov2023segment} for referring image segmentation (RIS) in the general domain, and fine-tunes it on MIMIC-ILS without further modification.
Like ROSALIA, most text-guided segmentation models in the medical domain couple a VLM with a segmentation module and predict the mask in a single shot~\citep{zhao2024biomedparse, huang2025towards}.
Such a design offers the user no view of how the mask was reached, and it is overly simplistic for lesion segmentation, which demands clinical expertise.

Recent work such as IBISAgent~\citep{jiang2026ibisagent} moves beyond a single
shot by having a VLM place point prompts on SAM over multiple turns, yet it
relies on points alone, carving the mask at a fine-grained level from the start
and thus requiring many steps to complete it. Radiologists in fact proceed through several levels of perception in sequence to delineate a lesion. They first survey the whole thorax to form a global impression, then narrow their attention to the approximate region where an abnormality is likely to reside, and only at the last stage refine the exact boundary at a fine-grained level.
Single-shot models collapse these levels into one step, and point-only refinement covers only the last.

\begin{figure*}[t]
\centering
\includegraphics[width=0.9\textwidth]{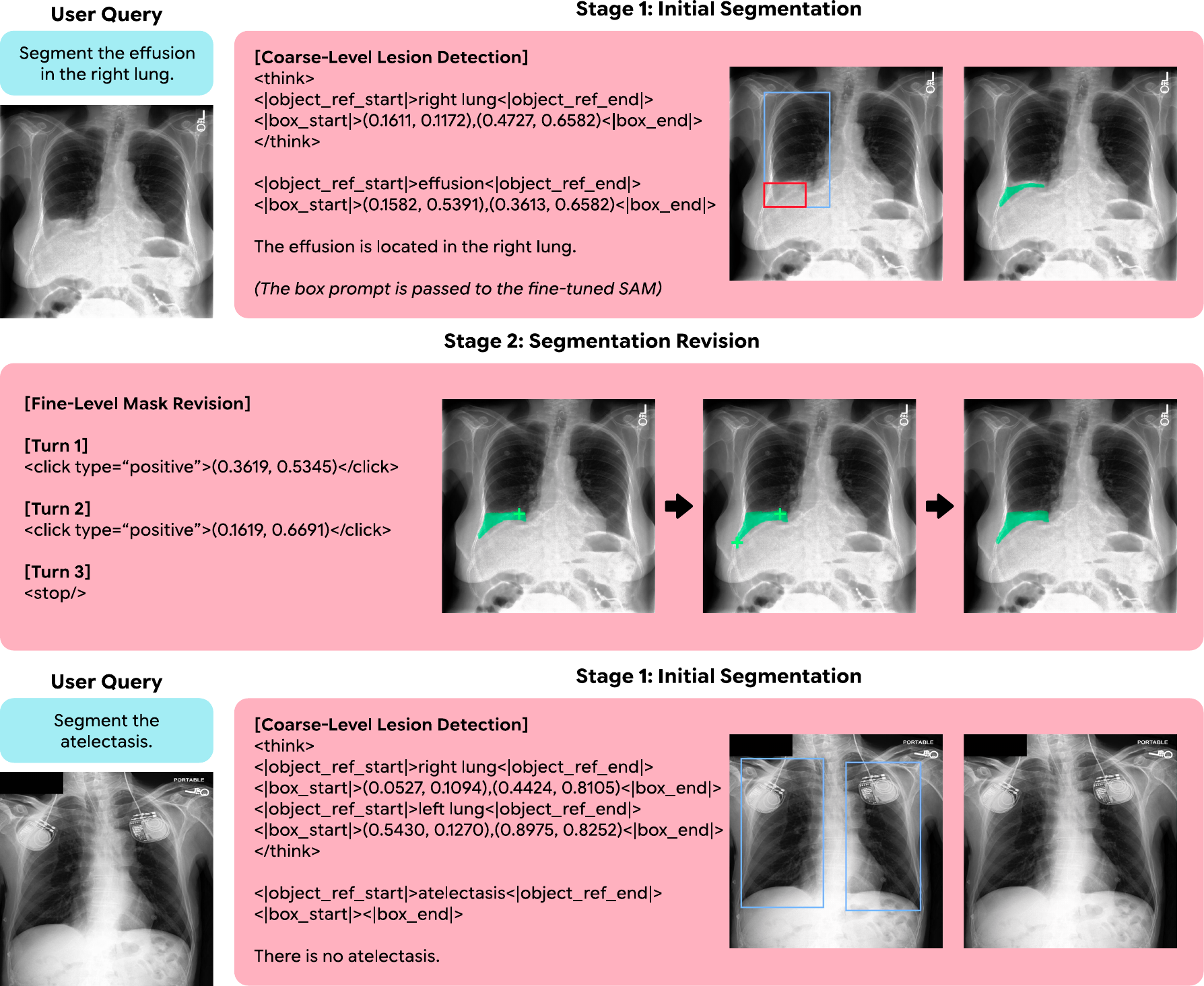}
\caption{Overview of \model. (Top) In Stage 1, \model grounds the lung regions, localizes the finding as a bounding box, and prompts a fine-tuned SAM with that box to obtain an initial mask.
In Stage 2, it inspects the mask and issues positive or negative point prompts that expand or shrink the predicted region, repeating this until the mask is satisfactory.
(Bottom) If the queried finding is absent, \model terminates at Stage 1 without producing a mask.}
\label{fig:teaser}
\end{figure*}

Following this observation, we develop \model, a workflow model for ILS in CXRs. \model performs segmentation as a sequence of two
stages that separate coarse-level from fine-level perception
(Figure~\ref{fig:teaser}). In the initial segmentation stage, the model reads
the cardiopulmonary field and first detects the lungs, then examines each of
them in turn to decide whether the queried finding is there, and draws a box
around the lesion, which it hands to a fine-tuned SAM as a prompt to obtain the
first mask. The subsequent segmentation revision stage inspects that mask and
corrects it through point prompts to the SAM, step by step, until the
boundary follows the lesion. Every intermediate decision is written out explicitly, so the user can trace
how the final mask was reached.

In our experiments, \model outperforms all compared models in segmentation quality on both internal and external evaluation, and an ablation study shows that each stage of the workflow, together with its training scheme, contributes to the gain. Trained on the same MIMIC-ILS cases as ROSALIA, \model improves segmentation by a clear margin, indicating that the gain does not come from additional lesion annotations.

\vspace{-0.5em}

Our contributions are summarized as follows:

\vspace{-0.5em}

\begin{itemize}[leftmargin=3.5mm]
    \item We present \model, a workflow model for CXR lesion
    segmentation, which decomposes the task into two stages following the order in which radiologists
    read an image.
    \item Trained on the same lesion annotations as ROSALIA, \model outperforms it
    by a clear margin on both internal and external evaluation, and an ablation
    shows that each component of the decomposition contributes to the gain.
    \item In a blinded reader study with three medical experts, masks from
    \model are ranked close to the ground truth and are preferred over those
    of ROSALIA about twice as often.
\end{itemize}

\vspace{-0.5em}

\section{Related Works}

\vspace{-0.5em}

\paragraph{Segmentation with VLMs and SAM.}
To segment a target described in free-form text, general-domain work couples a
VLM with SAM, and the resulting models differ mainly in how the VLM communicates
with the segmenter. LISA~\citep{lai2024lisa} feeds the hidden state of a special
\texttt{[SEG]} token to the SAM decoder, whereas later models have the VLM emit
geometric prompts that SAM consumes directly. Seg-Zero~\citep{liu2025seg}
predicts a box and points together and prompts SAM with them in a single step,
while SegAgent~\citep{zhu2025segagent} relies on point clicks alone, placing them
over multiple turns in imitation of a human annotator. In the medical domain,
IBISAgent~\citep{jiang2026ibisagent} follows the latter design, with a VLM
issuing point prompts across turns.

Both designs, however, have limitations. A single-step prompt cannot correct
the mask once it is produced, while a point-only procedure forgoes the box and
must carve out the mask click by click, taking IBISAgent 8.69 steps on average.
Moreover, IBISAgent is supervised on individual revision steps, and although it
later applies RL to its point prompts, it still optimizes each turn in
isolation, so the policy never learns how its clicks should work together
across the trajectory. \model combines the two designs, first localizing the
lesion with a box and then refining its boundary through successive point
prompts within at most four steps, and trains the revision policy with a reward
defined over the entire multi-turn trajectory.

\vspace{-0.5em}

\paragraph{Instruction-Guided Lesion Segmentation in Chest X-rays.}
Earlier models for text-guided lesion segmentation in CXRs are confined to a single target,
largely because annotated data is scarce~\citep{huang2024cross, li2023lvit}.
CheXlocalize~\citep{saporta2022benchmarking} provides radiologist-drawn masks for
CXR findings, but it is far too small to train on. To overcome this, MIMIC-ILS
introduced a large-scale lesion segmentation dataset built through an automated
framework, together with ROSALIA, a LISA-based model trained on
it~\citep{choi2026instruction, PhysioNet-mimic-cxr-ext-ils-1.0.0}. ROSALIA
accepts a text instruction such as ``Segment the opacity'' and covers seven
clinically important findings, namely cardiomegaly, pneumonia, atelectasis,
opacity, consolidation, edema, and effusion. Its masks, however, are often
imprecise, and because it predicts them in one shot, it neither follows the
coarse-to-fine workflow of a radiologist nor shows the user how the final
delineation was reached. \model addresses both: it covers the same findings
through this workflow, and because every intermediate decision is written out in
text, each of them remains inspectable by the user.

\vspace{-0.5em}

\section{Task Formulation}

\vspace{-0.5em}

We address instruction-guided lesion segmentation (ILS), the task introduced in
MIMIC-ILS. An instruction $Q = (c, \ell)$ names a target lesion $c$ and a location
$\ell$ at which to look, where $\ell$ ranges from the whole thorax (``Segment
the pleural effusion.'') through one lung (``\ldots in the left lung.'') to a
specific zone (``\ldots in the left lung base.''). Given a chest X-ray $I$ and
such an instruction, a model $f$ returns a binary mask over the image,
\begin{equation}
  M^{\text{pred}} = f(I, Q), \qquad
  M^{\text{pred}} \in \{0,1\}^{H \times W},
\end{equation}
covering the region the instruction refers to.

ILS is a medical counterpart of referring image segmentation (RIS), with one
difference that follows from how the queries arise. In RIS, the referring
expression describes an object the image is known to contain, so the target mask
is nonempty by construction, whereas a radiologist is as often asked whether a
finding is there at all. When $c$ is absent from $\ell$, the ground truth is the
empty mask, $M^{\text{GT}} = \mathbf{0}$. Whatever the instruction specifies,
then, the model must arrive at its own verdict on the region in question rather
than take the instruction as evidence that something is there.
 
\section{\model}

\subsection{Architecture Overview}

\paragraph{Notation.}
\model and a single promptable segmenter $F_{\text{seg}}$, built on SAM and shared by
both. The two policies interact with $F_{\text{seg}}$ through its geometric
prompts (boxes and points) rather than predicting the mask directly, and are applied in sequence.

Every region the model refers to is represented as a grounded box $g = (e, b)$,
where $e$ is a referring expression naming the region (e.g., right lung, pneumonia) and
$b = (x_1, y_1, x_2, y_2) \in [0,1]^4$ is its bounding box in normalized image
coordinates. We denote by $\mathcal{L}$ and $\mathcal{B}$ the sets of grounded
boxes for the lungs and for the lesion, respectively.

\paragraph{Stage 1: Initial segmentation.}
The first policy $\pi_{\text{init}}$ follows the order in which a radiologist
reads the image. It first grounds the lung regions to establish the anatomical
context in which the lesion should be sought, then judges whether the queried
lesion is present, and only then localizes it. These three decisions are emitted
as a single sequence,
\begin{equation}
  (\mathcal{L}, \mathcal{B}) \sim \pi_{\text{init}}(\cdot \mid I, Q),
\end{equation}
where the lung boxes $\mathcal{L}$ precede the lesion boxes $\mathcal{B}$. The scope of $\mathcal{L}$ follows the
instruction. A query about the thorax as a whole grounds both lungs,
$|\mathcal{L}| = 2$, whereas a query restricted to one side grounds only that
lung, $|\mathcal{L}| = 1$. When the lesion is absent, the policy emits no lesion
box, $\mathcal{B} = \emptyset$, and the instruction is answered with an empty
mask. Otherwise each lesion box prompts the segmenter separately, yielding a set of
initial masks
\begin{equation}
  \mathcal{M}_0 = \{\, F_{\text{seg}}(I, b) \mid b \in \mathcal{B} \,\}.
\end{equation}


\paragraph{Stage 2: Segmentation revision.}
Each mask in $\mathcal{M}_0$ is revised independently, and the union of the
revised masks forms the final prediction. Below we describe the procedure for
one such mask, denoted $M_0$. The second policy $\pi_{\text{rev}}$ refines $M_0$
over at most $T = 4$ steps. At step $t$, it observes the image overlaid with the
current mask and the point prompts
issued so far, denoted $o_t$, together with the history of preceding steps
$P_{<t}$, and produces an action $a_t$,
\begin{equation}
  a_t \sim \pi_{\text{rev}}(\cdot \mid I, Q, o_t, P_{<t}).
\end{equation}
The action space consists of point prompting and termination. A point prompt is
a pair $(s_t, p_t)$ with polarity $s_t \in \{+1, -1\}$ and relative coordinate
$p_t \in [0,1]^2$, where a positive point expands and a negative point shrinks
the predicted region. Since the box prompt confines the region the segmenter can
cover, we enlarge it to the smallest box enclosing both $b_{t-1}$ and $p_t$
whenever a positive point falls outside it, and leave it unchanged otherwise,
with $b_0$ set to the box that produced $M_0$. Executing the prompt yields the
next mask
\begin{equation}
  M_t = F_{\text{seg}}\!\left(I, b_t, \{(s_i, p_i)\}_{i \le t}, M_{t-1}\right),
\end{equation}
which is rendered as the observation $o_{t+1}$ and appended to the trajectory.
The loop terminates when the policy judges the mask to be sufficient or when
$t = T$. Because the entire revision proceeds within a single context, every
step is conditioned on the full history of the preceding ones. Applying this
procedure to every mask in $\mathcal{M}_0$ and taking the union of the results
gives the final prediction $M^{\text{pred}}$.

\subsection{Training Data}
\label{sec:training_data}
Training \model involves three components. The segmenter $F_{\text{seg}}$ is
fine-tuned for CXR lesions, and each of the two policies is trained through
supervised fine-tuning (SFT) followed by reinforcement learning (RL). All three
are trained on data constructed automatically from
MIMIC-ILS~\citep{choi2026instruction, PhysioNet-mimic-cxr-ext-ils-1.0.0}, since
every field is obtained from the masks and metadata the dataset already
provides, so building it requires no additional human annotation. We divide MIMIC-ILS into disjoint splits for SFT and RL ($D_{\text{SFT}}$ and $D_{\text{RL}}$), both of which are also disjoint from the evaluation split $D_{\text{eval}}$. For RL and evaluation we draw both sets from the cases that medical experts in
the CheXpercept study~\citep{choi2026chexpercept} identified as the most
reliable within MIMIC-ILS in terms of mask quality.

\subsection{SAM Training}

Within \model, $F_{\text{seg}}$ is invoked by the policies, so the segmenter
must be able to respond to prompts that point at lesions. Neither the original
SAM nor its medical variants such as MedSAM are trained on CXR lesions, so we
obtain $F_{\text{seg}}$ by fine-tuning SAM3~\citep{carion2026sam} on the SFT split $D_{\text{SFT}}$. From each
ground-truth mask $M^{\text{GT}}$ we derive the prompts used during training, sampling a
box $b$ that encloses the annotated region together with points $(s, p)$ drawn
from inside and outside it, so that the prompt configuration seen during
training matches the one the policies produce at inference time.

\begin{figure*}[t]
\centering
\includegraphics[width=\textwidth]{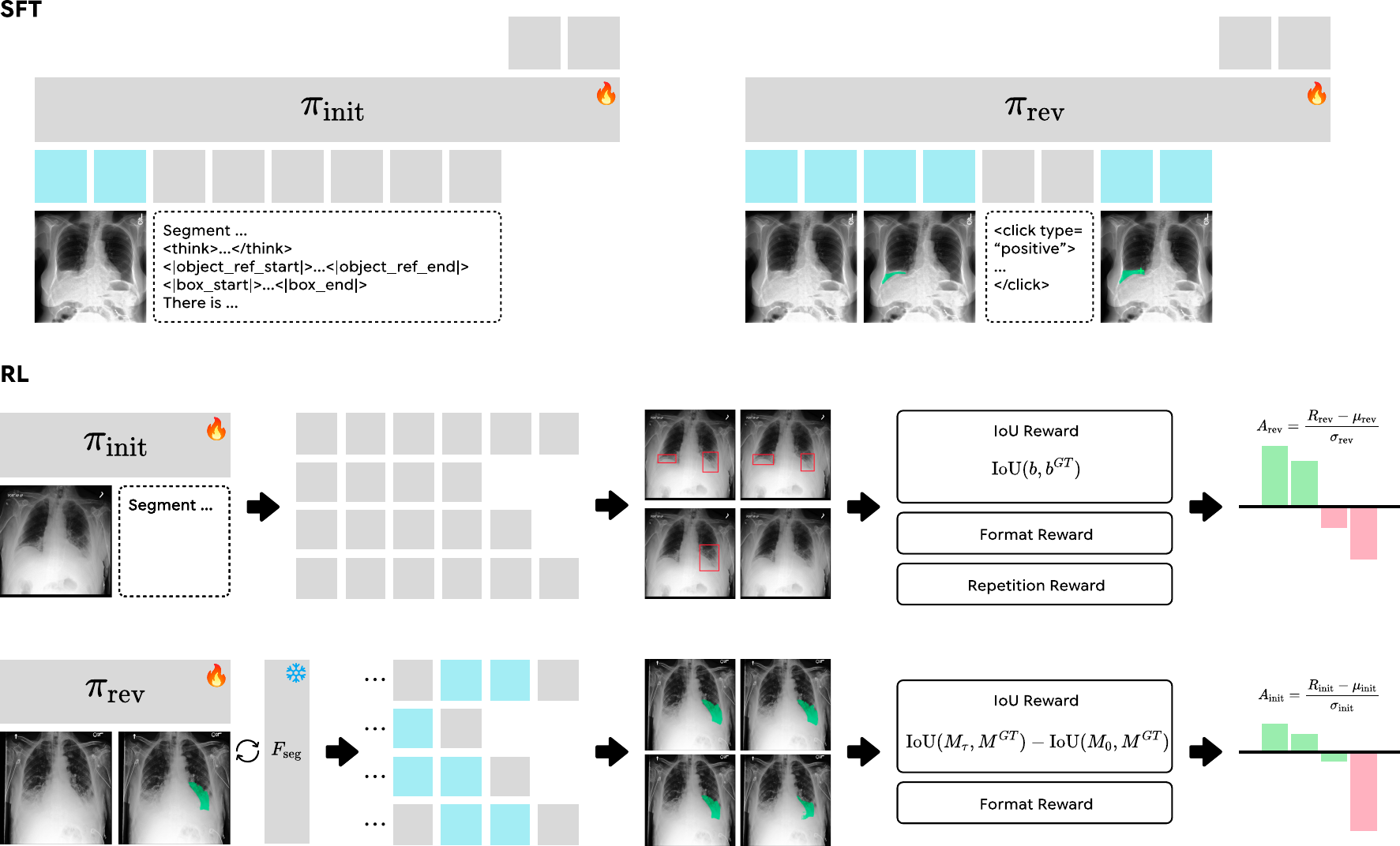}
\caption{Training the two policies. We first apply SFT to
$\pi_{\text{init}}$ and $\pi_{\text{rev}}$ on automatically constructed initial
segmentation and segmentation revision scenarios, and then train each policy
further with GRPO. $A_{\text{init}}$ and $A_{\text{rev}}$ are the advantages
obtained by normalizing $R_{\text{init}}$ and $R_{\text{rev}}$ within each group
of sampled generations. Text tokens are shown in gray and image tokens in light
blue.}
\label{fig:training}
\end{figure*}

\subsection{VLM Training}

The two policies are trained separately, and executed in sequence at
inference time. SFT alone optimizes only the token-level likelihood of the
target sequence. The model learns to imitate the format of the target sequence, but receives no
direct feedback on whether the decisions it contains are correct. We therefore continue with RL in both stages, using rewards that
measure the quality of the outcome. Throughout, $F_{\text{seg}}$ is kept frozen,
so the policies learn to steer a fixed segmenter. Prompt templates and example SFT targets are deferred to Appendix~\ref{sec:examples}.

\subsubsection{Initial Segmentation Policy}

Each SFT example for $\pi_{\text{init}}$ pairs an image and an instruction with the sequence of decisions the stage requires, the lung boxes, the presence decision, and the lesion boxes, written
out in the format shown in Figure~\ref{fig:teaser}. Lung
boxes are obtained by running an off-the-shelf lung segmentation model~\citep{seibold2023accurate, cosarinsky2025chexmask}, while
the presence label and the lesion boxes are derived directly from MIMIC-ILS.

The RL reward combines a localization term with two auxiliary terms,
\begin{equation}
  R_{\text{init}} = R_{\text{box}} + R_{\text{format}} + R_{\text{rep}},
  \qquad R_{\text{box}} = \mathrm{IoU}(b, b^{\text{GT}}).
\end{equation}
For a negative case, $R_{\text{box}}$ is instead binary and depends only on
whether the model emits $\mathcal{B} = \emptyset$. The other two terms act on
the form of the output rather than its content. $R_{\text{format}}$ checks that
the generated text follows the expected structure, so that the grounded boxes
can be parsed from it, and $R_{\text{rep}}$ penalizes emitting the same box more
than once.

\subsubsection{Segmentation Revision Policy}

SFT examples for $\pi_{\text{rev}}$ are revision trajectories constructed by
simulating the process the policy will later perform. We perturb the
ground-truth box $b^{\text{GT}}$ and feed it to $F_{\text{seg}}$, which yields
an imperfect mask $\tilde{M}$. Starting from $\tilde{M}$, we place clicks
following the simulated-user protocol standard in interactive
segmentation~\citep{xu2016deep, sofiiuk2022reviving, kirillov2023segment}, which
selects at each step the point that brings the resulting mask closest to
$M^{\text{GT}}$. One trajectory is thus
\begin{equation}
  \left(M_0, a_1, M_1, a_2, \ldots\right), \qquad M_0 = \tilde{M},
\end{equation}
where a point prompt $(s_t, p_t)$ turns $M_{t-1}$ into $M_t$, and the trajectory ends
either at the step limit or when the policy issues the termination action. Each
$M_t$ enters the example as the observation $o_{t+1}$, rendered by overlaying
the mask and the points issued so far on $I$, exactly as at inference time.

The RL reward pairs the gain in IoU over the trajectory the model itself
produces with the same format term as before,
\begin{equation}
  R_{\text{rev}} = R_{\text{gain}} + R_{\text{format}},
  \qquad R_{\text{gain}} = \mathrm{IoU}(M_{\tau}, M^{\text{GT}})
  - \mathrm{IoU}(M_0, M^{\text{GT}}),
\end{equation}
where $\tau$ is the step at which the policy terminates and $M_0 = \tilde{M}$ is
the mask the trajectory starts from. Rewarding the net gain lets the model stop
as soon as further revision would no longer help.

\vspace{-0.5em}

\section{Experiments}

\vspace{-0.5em}

\subsection{Implementation Details}
$F_{\text{seg}}$ is built on SAM3, fine-tuned with the image encoder unfrozen.
Both policies are built on
MedGemma-1.5-4B-it~\citep{sellergren2026medgemma} and trained on eight
A100-80GB GPUs with LoRA~\citep{hu2021lora} ($r{=}128$, $\alpha{=}256$) on all
linear layers and the vision tower unfrozen, at an effective batch size of $256$
for $\pi_{\text{init}}$ and $64$ for $\pi_{\text{rev}}$ during supervised
fine-tuning. RL then uses GRPO~\citep{shao2024deepseekmath} with $8$ generations
per prompt at temperature $1.0$, and the rollouts of $\pi_{\text{rev}}$ run with
$F_{\text{seg}}$ in the loop for at most four clicks per episode. Before that,
$\pi_{\text{rev}}$ is warm-started from $\pi_{\text{init}}$ at its SFT
checkpoint, so that it inherits grounding ability without also inheriting a
habit of always committing to a box. The remaining hyperparameters are given in
the Appendix~\ref{sec:training_details}.

\vspace{-0.5em}

\subsection{Experimental Setup}

\paragraph{Evaluation set.}
As described in Sec.~\ref{sec:training_data}, evaluation is carried out on part
of the CheXpercept subset of MIMIC-ILS, the cases whose masks physicians judged
most reliable. This portion is disjoint from all training data, so that every
evaluation image is unseen. Even within this subset, human error may remain, so
we apply an additional filter. We obtain further evidence on whether a lesion is
present and where it lies from two more sources, Chest
ImaGenome~\citep{wu2021chest, PhysioNet-chest-imagenome-1.0.0} and an LLM parse
of the radiology report accompanying the X-ray, and rebuild the evaluation set
from the cases on which all of them agree. The resulting set queries the same
studies at three levels of location specificity: chest, naming the finding alone
(``Segment the edema.''; 382 positives, 444 negatives); lung, adding the side
(``\ldots in the left lung.''; 429, 1,050); and zone, naming the regions
directly (``\ldots in the left mid zone and left lung base.''; 330, 3,286).
Appendices~\ref{sec:agreement} and~\ref{sec:evaluation} give the agreement
criteria and the per-finding counts.

We additionally evaluate on CheXlocalize~\citep{saporta2022benchmarking}, an external dataset whose images
come from CheXpert~\citep{irvin2019chexpert} and whose masks are drawn from scratch
by radiologists. We use the 775 positive and 1,643 negative cases over the six findings
that CheXlocalize shares with the set of lesions \model can segment, and query them at the chest level only (e.g., ``Segment the consolidation''). CheXlocalize also
provides segmentations drawn by a second, independent pair of radiologists,
which we score through the same pipeline as the models and report as a human
benchmark.

\vspace{-0.5em}

\paragraph{Baselines and metrics.}
We compare our model against two groups of models. The first consists of general-domain models, LISA~\citep{lai2024lisa}, PixelLM~\citep{ren2024pixellm}, and Text4Seg~\citep{lan2025text4seg}, which take a
free-form instruction and return a mask. The
second covers models developed for the medical domain, BiomedParse~\citep{zhao2024biomedparse}, RecLMIS~\citep{huang2024cross},
IMIS-Net~\citep{cheng2025interactive} and IBISAgent~\citep{jiang2026ibisagent}%
\renewcommand{\thefootnote}{\ensuremath{\dagger}}%
\footnote{\label{fn:ibis}IBISAgent is an intermediate training checkpoint, rather than the final model, has been publicly released, and the model cannot be reliably retrained, as the released code does not cover the full training pipeline in sufficient detail and its data construction relies on manual annotations by physicians.}%
\renewcommand{\thefootnote}{\arabic{footnote}}\addtocounter{footnote}{-1}%
. We also evaluate fine-tuned ROSALIA~\citep{choi2026instruction} on the same MIMIC-ILS cases ($D_{\text{SFT}}$ and $D_{\text{RL}}$). We report gIoU, the
mean per-sample IoU, and cIoU, the ratio of total intersection to total union.
F1 measures how well a model separates positive from negative cases, treating
a returned mask as positive and an empty output as negative. 


\vspace{-0.5em}

\subsection{Main Results}

\vspace{-0.5em}

\paragraph{Comparison with prior models.}

\model records the highest gIoU and cIoU of all models we compare against,
on both CheXpercept and CheXlocalize. The margin is largest over the baselines
outside the ROSALIA family, but it also holds over ROSALIA, the only prior model
trained for this task. For a fair comparison, we retrain ROSALIA on exactly the
cases \model uses, so that both models receive the same lesion supervision
and the gain cannot be attributed to additional annotations. \model raises
gIoU on CheXpercept from 0.621 to 0.703 and F1 from 0.945 to 0.961. The same
pattern holds on CheXlocalize, where gIoU rises from 0.233 to 0.283 and cIoU
from 0.289 to 0.331, approaching the human benchmark at 0.314 and 0.339, while
F1 remains on par (0.632 vs.\ 0.634). Since these masks were drawn independently
by radiologists, the gain carries over to an annotation protocol our training
data never saw.


\vspace{-0.5em}

\paragraph{Effect of each stage and its training phase.}
We examine how much SFT and RL contribute to each stage
(Table~\ref{tab:ablation}). Each addition improves or maintains gIoU and
consistently improves cIoU on both benchmarks. Most of the gain comes from
Stage~1. SFT alone reaches a gIoU of 0.668 on CheXpercept, above ROSALIA's
0.621, though its F1 of 0.932 falls short of ROSALIA's 0.945. RL raises gIoU
to 0.693 and F1 to 0.961, suggesting that the outcome-based reward sharpens the
presence decision and box placement that SFT only imitates. Stage~2 adds a
further 0.010 gIoU on CheXpercept and 0.003 on CheXlocalize, as it only makes
local corrections and can neither recover a lesion Stage~1 missed nor change a
presence decision.

\newcommand{\yes}{\checkmark}
\begin{table}[t]
\centering
\caption{Comparison with prior models. On
CheXpercept the three question levels are averaged with equal weight, and on
CheXlocalize every case is queried at the chest level. The radiologists behind
the human benchmark were told which finding to draw, so F1 is not defined for
them. The best score among the models is shown in \textbf{bold} and the second best is \underline{underlined}. 
}
\label{tab:main}
\vspace{2mm}
\small
\setlength{\tabcolsep}{9pt}
\begin{tabular}{lcccccc}
\toprule
\multirow{2}{*}{Model}
& \multicolumn{3}{c}{\textbf{CheXpercept}}
& \multicolumn{3}{c}{\textbf{CheXlocalize}} \\
\cmidrule(lr){2-4}\cmidrule(l){5-7}
& gIoU & cIoU & F1 & gIoU & cIoU & F1 \\
\midrule
\textit{Human benchmark} & --- & --- & --- & \textit{0.314} & \textit{0.339} & --- \\
\midrule
LISA-7B & 0.014 & 0.024 & 0.401 & 0.082 & 0.119 & 0.491 \\
PixelLM-7B & 0.031 & 0.032 & 0.416 & 0.125 & 0.125 & 0.485 \\
Text4Seg & 0.035 & 0.038 & 0.402 & 0.119 & 0.132 & 0.475 \\
IMIS-Net & 0.080 & 0.087 & 0.413 & 0.005 & 0.007 & 0.423 \\
IBISAgent-7B$^{\dagger}$ & 0.135 & 0.131 & 0.416 & 0.081 & 0.082 & 0.484 \\
BiomedParse & 0.257 & 0.254 & 0.413 & 0.105 & 0.108 & 0.483 \\
RecLMIS & 0.302 & 0.305 & 0.419 & 0.138 & 0.143 & 0.488 \\
ROSALIA & \underline{0.621} & \underline{0.688} & \underline{0.945} & \underline{0.233} & \underline{0.289} & \textbf{0.634} \\

\midrule
\rowcolor{oursgray}
\model (ours) & \textbf{0.703} & \textbf{0.731} & \textbf{0.961} & \textbf{0.283} & \textbf{0.331} & \underline{0.632} \\
\bottomrule
\end{tabular}
\end{table}
\vspace{-1em}
\begin{table}[t]
\centering
\caption{Contribution of each stage and training. Every component is added on top of
the one before it. Stage~2
refines boxes Stage~1 has already emitted and never changes a presence decision,
so it inherits Stage~1's F1. ROSALIA is shown for reference and
columns are as in Table~\ref{tab:main}. The best score is shown in \textbf{bold} and the second best is \underline{underlined}.} 
\label{tab:ablation}
\vspace{2mm}
\small
\setlength{\tabcolsep}{6pt}
\begin{tabular}{lcccccccccc}
\toprule
\multirow{2}{*}{Model}
& \multicolumn{2}{c}{\textbf{Stage 1}} & \multicolumn{2}{c}{\textbf{Stage 2}}
& \multicolumn{3}{c}{\textbf{CheXpercept}}
& \multicolumn{3}{c}{\textbf{CheXlocalize}} \\
\cmidrule(lr){2-3}\cmidrule(lr){4-5}\cmidrule(lr){6-8}\cmidrule(l){9-11}
& SFT & RL & SFT & RL & gIoU & cIoU & F1 & gIoU & cIoU & F1 \\
\midrule
ROSALIA &      &      &      &      & 0.621 & 0.688 & 0.945 & 0.233 & 0.289 & \textbf{0.634} \\
\midrule
\multirow{4}{*}{\model}
        & \yes &      &      &      & 0.668 & 0.709 & 0.932 & 0.268 & 0.325 & 0.631 \\
        & \yes & \yes &      &      & 0.693 & 0.718 & \textbf{0.961} & 0.280 & 0.327 & \underline{0.632} \\
        & \yes & \yes & \yes &      & \underline{0.699} & \underline{0.726} & \textbf{0.961} & \underline{0.280} & 0.329 & \underline{0.632} \\
        & \yes & \yes & \yes & \yes & \textbf{0.703} & \textbf{0.731} & \textbf{0.961} & \textbf{0.283} & \textbf{0.331} & \underline{0.632} \\
\bottomrule
\end{tabular}
\end{table}

\begin{figure*}[t]
\centering
\includegraphics[width=\textwidth]{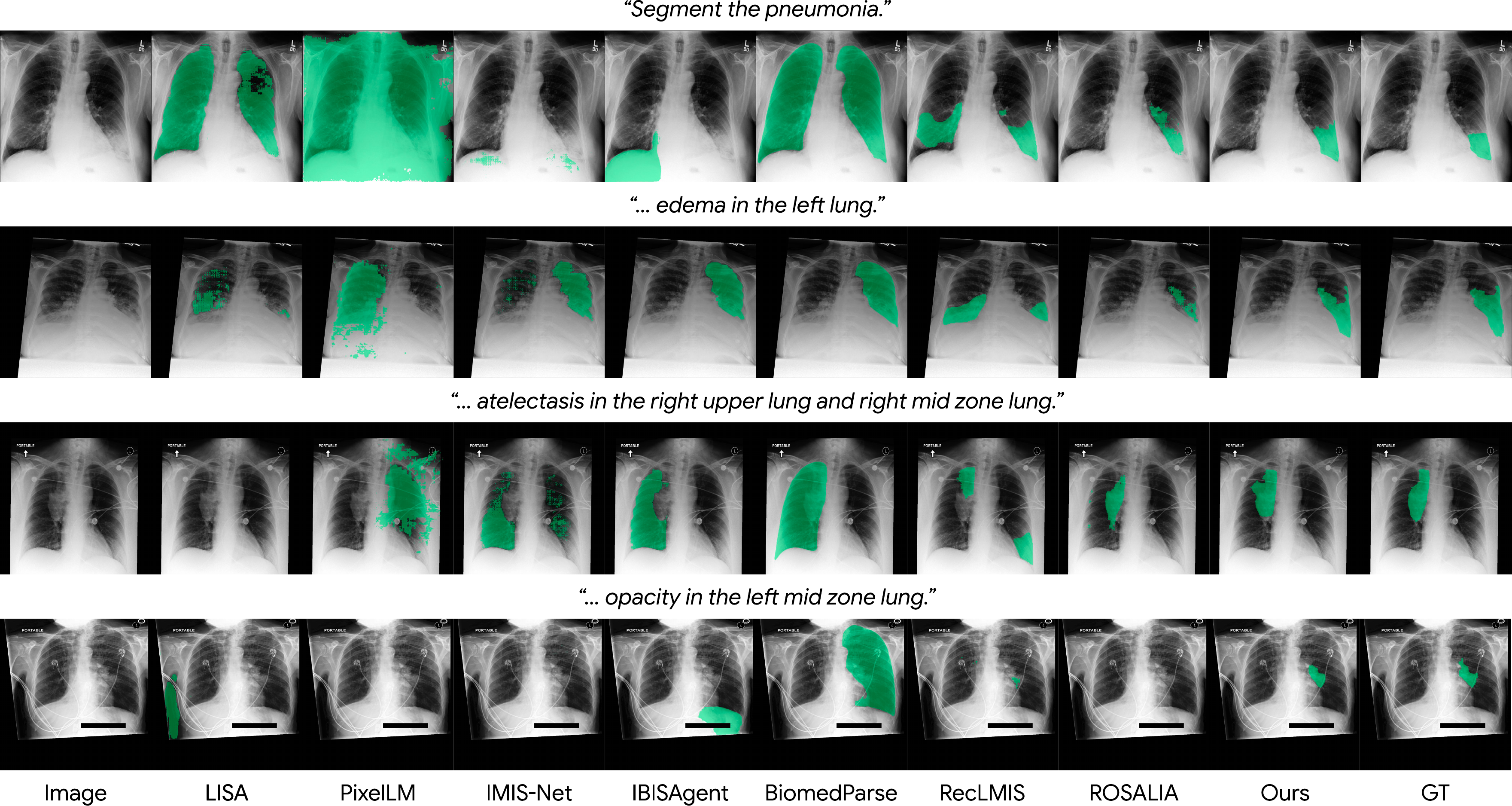}
\caption{Qualitative comparison of \model against baseline models.}
\label{fig:qualitative}
\end{figure*}

\subsection{Qualitative Results}

Figure~\ref{fig:qualitative} shows how the predicted masks differ across models.
Apart from the ROSALIA family, the baselines do not follow the instruction in
any meaningful sense and have little ability to delineate a lesion, returning
regions that bear no relation to the finding that was asked for. ROSALIA
responds to the instruction but its masks are imprecise. It misses small lesions
entirely, and where it does respond, the predicted boundary fails to track the
lesion contour and is accompanied by scattered spurious fragments. \model,
by contrast, recovers even small lesions lying close to the heart, and its
boundaries follow the ground truth closely with far fewer stray regions. Additional qualitative examples are provided in Appendix~\ref{sec:additional}.

\section{Further Analyses}

\subsection{Mask Noise}
\label{sec:mask_noise}
Physicians tend to delineate a finding as a single smooth, continuous region unless it appears in clearly separate locations.
\model's masks are visibly closer to this than ROSALIA's, yet IoU cannot capture the difference.
We therefore count connected components relative to the GT mask for the same case,
\begin{equation}
  \Delta C = C(M^{\text{pred}}) - C(M^{\text{GT}}),
\end{equation}
where $C(\cdot)$ is the number of connected components in a mask.
$\Delta C$ thus measures how many more components a model produces than the GT.
A positive value means the prediction breaks the finding into excess fragments, and a negative value means it yields fewer components than the GT.
Additionally, we count specks, which we define as components smaller than $0.1\%$ of the image area.
As shown in Table~\ref{tab:noise}, ROSALIA produces $5.63$ more components than the GT on average, whereas \model produces $1.71$.
Most of this excess consists of specks, as ROSALIA leaves $5.57$ specks per case against \model's $1.69$.
Of these, $3.43$ and $1.06$, respectively, touch no part of the GT.
These specks account for $2.6\%$ of ROSALIA's mask area but only $0.14\%$ of \model's.
\model therefore produces fewer small mask fragments that do not belong to the lesion.

\begin{table}[t]
\centering
\caption{Mask fragmentation analysis over all 1{,}141 positive cases in the CheXpercept evaluation set. $\Delta C$ and specks are defined in Section~\ref{sec:mask_noise}. \textit{Outside GT} counts the specks that do not overlap the GT mask, and \textit{Mask area} is the fraction of the predicted mask area occupied by specks.}
\label{tab:noise}
\vspace{2mm}
\small
\setlength{\tabcolsep}{6pt}
\begin{tabular}{lcccc}
\toprule
\multirow{2}{*}{Model} & \multirow{2}{*}{$\Delta C$}
& \multicolumn{3}{c}{\textbf{Specks}} \\
\cmidrule(lr){3-5}
& & count $\downarrow$ & outside GT $\downarrow$
& mask area \% $\downarrow$ \\
\midrule
\textit{GT} & \textit{---} & \textit{0.01} & \textit{---} & \textit{0.01} \\
\rowcolor{oursgray}
\model (ours) & $+1.71$ & 1.69 & 1.06
& 0.14 \\
ROSALIA & $+5.63$ & 5.57 & 3.43 & 2.60 \\
\bottomrule
\end{tabular}
\vspace{-1em}
\end{table}

\subsection{Expert Preference Study}

We additionally ran a reader study in which medical experts ranked the mask quality of the ground truth, \model, and ROSALIA, which we include as the previous state of the art. For every case the reader
saw the three masks in a counterbalanced random order, without any indication of their source,
and ranked them by how well each delineated the queried finding, with ties
allowed when two masks looked comparable. Three physicians each reviewed the same
95 cases, drawn so that every finding and question level is equally represented.
Cardiomegaly contributes five cases at the chest level, and each of the remaining six findings contributes five cases at each of the three levels (e.g., ``Segment the atelectasis'', ``... in the right lung'', and ``... in the right lung base'').

Table~\ref{tab:reader} reports the result. We fit two tie-aware extensions of
the Bradley--Terry model~\citep{Bradley1952RankAO}, Davidson~\citep{Davidson1970OnET} and Rao--Kupper~\citep{Rao01031967}, which differ in how they
treat a tie, and report the log-strength each assigns to the three masks.
\model sits close to the ground truth, a difference of $0.141$ under
Davidson, so that a reader who separates the two prefers the ground truth
$1.15$ times as often, and their intervals overlap over most of their range.
ROSALIA falls well short of both \model is preferred over it $2.01$ times
as often, a difference of $0.696$, and the ground truth by more still, with
ROSALIA's interval overlapping neither of theirs. More details on the human study are provided in Appendix~\ref{sec:human_study}.

\begin{table}[t]
\centering
\caption{Expert preference study. Three medical experts ranked three
blinded masks on 95 balanced positive cases, giving 285 completed rankings.
Log-strengths come from two tie-aware extensions of the Bradley--Terry model,
Davidson and Rao--Kupper, which differ in how
they treat a tie but agree here. Values are centred to sum to zero, with $95\%$
bootstrap intervals over cases, and a difference of $d$ means the stronger of
the two masks is preferred $e^{d}$ times as often when the reader does not tie.}
\label{tab:reader}
\vspace{2mm}
\small
\setlength{\tabcolsep}{7pt}
\begin{tabular}{lccc}
\toprule
\multirow{2}{*}{Model} & \multicolumn{2}{c}{\textbf{Log-strength} $\uparrow$}
& \multirow{2}{*}{Mean rank $\downarrow$} \\
\cmidrule(lr){2-3}
& Davidson & Rao--Kupper & \\
\midrule
GT & 0.326 {\scriptsize [0.17, 0.48]}
& {0.294} {\scriptsize [0.16, 0.44]} & 1.79 \\
\rowcolor{oursgray}
\model (ours) & 0.185 {\scriptsize [0.01, 0.38]}
& 0.170 {\scriptsize [0.01, 0.33]} & 1.88 \\
ROSALIA & $-0.511$ {\scriptsize [$-0.72$, $-0.33$]}
& $-0.464$ {\scriptsize [$-0.64$, $-0.30$]} & 2.33 \\
\bottomrule
\end{tabular}
\vspace{-1em}
\end{table}

\section{Conclusion}

We introduced \model, a workflow model for CXR lesion segmentation.
\model mirrors the way a radiologist reads an image, separating coarse
localization from fine-grained delineation and carrying them out in sequence
rather than mapping an image to a mask in one step. \model substantially outperforms prior models and surpasses ROSALIA, the previous state of the art, even when the two are trained on the same lesion annotations. In future work, the intermediate steps and masks \model produces could serve as a foundation for other CXR tasks.



\bibliographystyle{plainnat}
\bibliography{paper}

\clearpage
\appendix
\newpage
\section{Training}
\label{sec:training}

\subsection{Data Curation}
Every split is built from two sources. MIMIC-ILS~\citep{choi2026instruction,
PhysioNet-mimic-cxr-ext-ils-1.0.0} supplies the images, the lesion masks and the
report-derived labels, and CheXpercept~\citep{choi2026chexpercept} supplies the
subset of those cases whose masks medical experts reviewed and judged reliable.
MIMIC-ILS is large enough to train on, while CheXpercept is small and
trustworthy enough to reinforce against and to score on. We therefore hold all
$2{,}100$ CheXpercept radiographs out of the supervised split and divide them in
half, one half for RL and one for evaluation. What follows describes the splits
themselves. Each stage builds its own prompts, boxes and masks from the studies
its split contains, so one study gives rise to many samples, and a question
asked at three levels of location specificity gives rise to three.

\begin{table}[h]
\centering
\caption{How the splits divide the two sources. CheXpercept is a
curated subset of the MIMIC-ILS training split, so holding it out leaves the
supervised split disjoint from the other two by construction. Counts are
radiographs; MIMIC-ILS carries one frontal radiograph per study, so they are
also study counts.}
\label{tab:splits}
\vspace{2mm}
\small
\setlength{\tabcolsep}{4pt}
\begin{tabular}{llcl}
\toprule
Split & Drawn from & Radiographs & Selection \\
\midrule
\multirow{2}{*}{$D_{\text{SFT}}$}
 & MIMIC-ILS train $\setminus$ CheXpercept & $185{,}488$
 & everything the review did not cover \\
 & MIMIC-ILS val & $1{,}508$
 & MIMIC-ILS's own split, unchanged \\
\midrule
$D_{\text{RL}}$ & MIMIC-ILS train $\cap$ CheXpercept & $1{,}296$
 & $70\%$ of the agreed set, by patient \\
$D_{\text{eval}}$ & MIMIC-ILS train $\cap$ CheXpercept & $560$
 & the remaining $30\%$ \\
\midrule
\multicolumn{2}{l}{\textit{MIMIC-ILS train, before the exclusion}}
& $187{,}588$ & \\
\multicolumn{2}{l}{\textit{CheXpercept, all of it inside MIMIC-ILS train}}
& $2{,}100$ & \\
\multicolumn{2}{l}{\textit{of which the three sources agree on}}
& $1{,}856$ & \\
\bottomrule
\end{tabular}
\end{table}

\paragraph{$D_{\text{SFT}}$.} The MIMIC-ILS training split with the CheXpercept
radiographs removed, together with MIMIC-ILS's own validation split. CheXpercept
was drawn entirely from the training split, all $2{,}100$ of its radiographs
sitting there and none in validation or test, so removing them leaves
$185{,}488$ of the original $187{,}588$ and leaves the $1{,}508$ validation
radiographs untouched.

\paragraph{$D_{\text{RL}}$ and $D_{\text{eval}}$.} Both come from CheXpercept,
and both are filtered by the three-source agreement of
Section~\ref{sec:agreement} before being divided, which leaves $1{,}856$ of the
$2{,}100$ radiographs.

The division is by patient rather than by radiograph. Those $1{,}856$
radiographs come from $1{,}526$ patients, and $226$ of them contribute more than
one, covering $556$ radiographs in all, so a split on radiograph identity would
put the same chest on both sides. Patients are assigned greedily to whichever
side is furthest below its quota on the scarcest stratum that patient carries,
because matching the overall target ratios is not enough on its own: the
evaluation's power to test per-side discrimination rests on the $85$ unilateral
(radiograph, finding) pairs and on the per-finding positives, both of which a
ratio-matched split can strand almost entirely on one side. The result is
$1{,}296$ radiographs for $D_{\text{RL}}$ and $560$ for $D_{\text{eval}}$, with
no patient in both.

\subsection{Three-Source Agreement}
\label{sec:agreement}
CheXpercept's labels can be wrong, and an evaluation set inherits whatever it
gets wrong, so we check each of them against two further sources and keep only
what all three support. All three record both whether a finding is present and
where it is, which is what makes the check possible; they differ in how they
record it.

\begin{itemize}[leftmargin=1.5em,itemsep=2pt,topsep=3pt]
\item \textbf{CheXpercept} gives the list of zones a finding occupies.
\item \textbf{The radiology report} gives free text, which we parse with
      MedGemma into a present or absent call per finding and region.
\item \textbf{Chest ImaGenome} gives a yes or no per finding and region in its
      silver scene graphs.
\end{itemize}

Each is then reduced to the same verdict, present or absent for one
(radiograph, finding, region). A triple is admitted only when the three verdicts
are identical, and takes that verdict as its label; anything else, including a
missing source, is dropped. $1{,}856$ of the $2{,}100$ radiographs keep at least
one admitted triple, and those are the radiographs $D_{\text{RL}}$ and
$D_{\text{eval}}$ are drawn from.

\subsection{Building the Training Examples}
\label{sec:examples}
Every example is automatically generated from what a split already carries, the
radiograph, the lesion mask and the report-derived label.
Table~\ref{tab:examples} reports how many examples this yields for each stage
and phase.

\begin{table}[h]
\centering
\caption{Examples built from each split. The unit differs by stage:
$\pi_{\text{init}}$ is trained one instruction at a time, while one
$\pi_{\text{rev}}$ example is a whole revision trajectory of up to four turns.}
\label{tab:examples}
\vspace{2mm}
\small
\begin{tabular}{llll}
\toprule
Stage & Split & Unit & Examples \\
\midrule
$\pi_{\text{init}}$ SFT & $D_{\text{SFT}}$ & instruction
 & $1{,}031{,}506$ train / $8{,}246$ val \\
$\pi_{\text{init}}$ RL & $D_{\text{RL}}$ & instruction
 & $13{,}488$ ($2{,}605$ positive, $10{,}883$ negative) \\
$\pi_{\text{rev}}$ SFT & $D_{\text{SFT}}$ & trajectory
 & $469{,}011$ train / $4{,}423$ val \\
$\pi_{\text{rev}}$ RL & $D_{\text{RL}}$ & starting state
 & $58{,}320$ \\
\bottomrule
\end{tabular}
\end{table}

\paragraph{Initial segmentation.}
A training example for this stage runs from a chest X-ray and a user question to
the bounding boxes the model is to emit. Questions are generated by filling
templates with the (finding, region) labels that accompany each image in
$D_{\text{SFT}}$, which keeps the phrasing varied while the underlying query
stays grounded in the annotation. The target answer begins with the lungs. We
obtain their boxes by running CXAS and CheXmask-U to extract lung masks and taking
the tightest box around each. Which lungs appear follows the question, so a question about
the thorax as a whole grounds both lungs and a question restricted to one side
grounds only that one. The lesion boxes follow immediately, derived from the
MIMIC-ILS mask for the queried finding. 

\begin{figure*}[h]
\centering
\includegraphics[width=\textwidth]{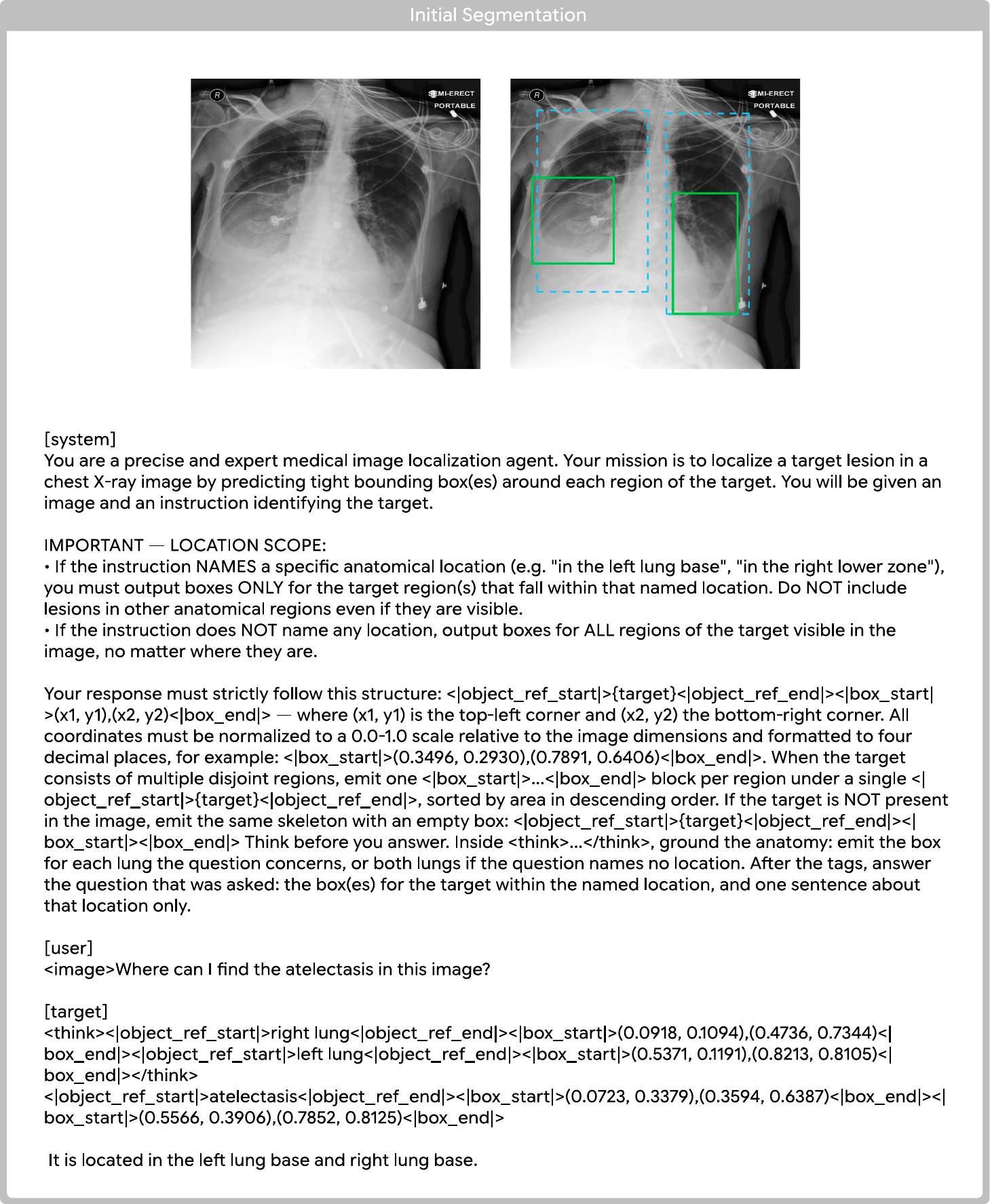}
\caption{Example of SFT data for initial segmentation. The image shown at the
top is what the \texttt{<image>} placeholder stands for; the box overlay is
drawn here for illustration and is not part of the input.}
\label{fig:stg1}
\end{figure*}

Every element of the answer is wrapped
in markers such as \texttt{<think>} and \texttt{<|box\_start|>} so that it can be
parsed from the generated text. Figure~\ref{fig:stg1} gives the full example. RL needs only the prompt
side of this, since the model generates its own answer and is scored on the
outcome rather than on matching a target.

\begin{figure*}[h]
\centering
\includegraphics[width=\textwidth]{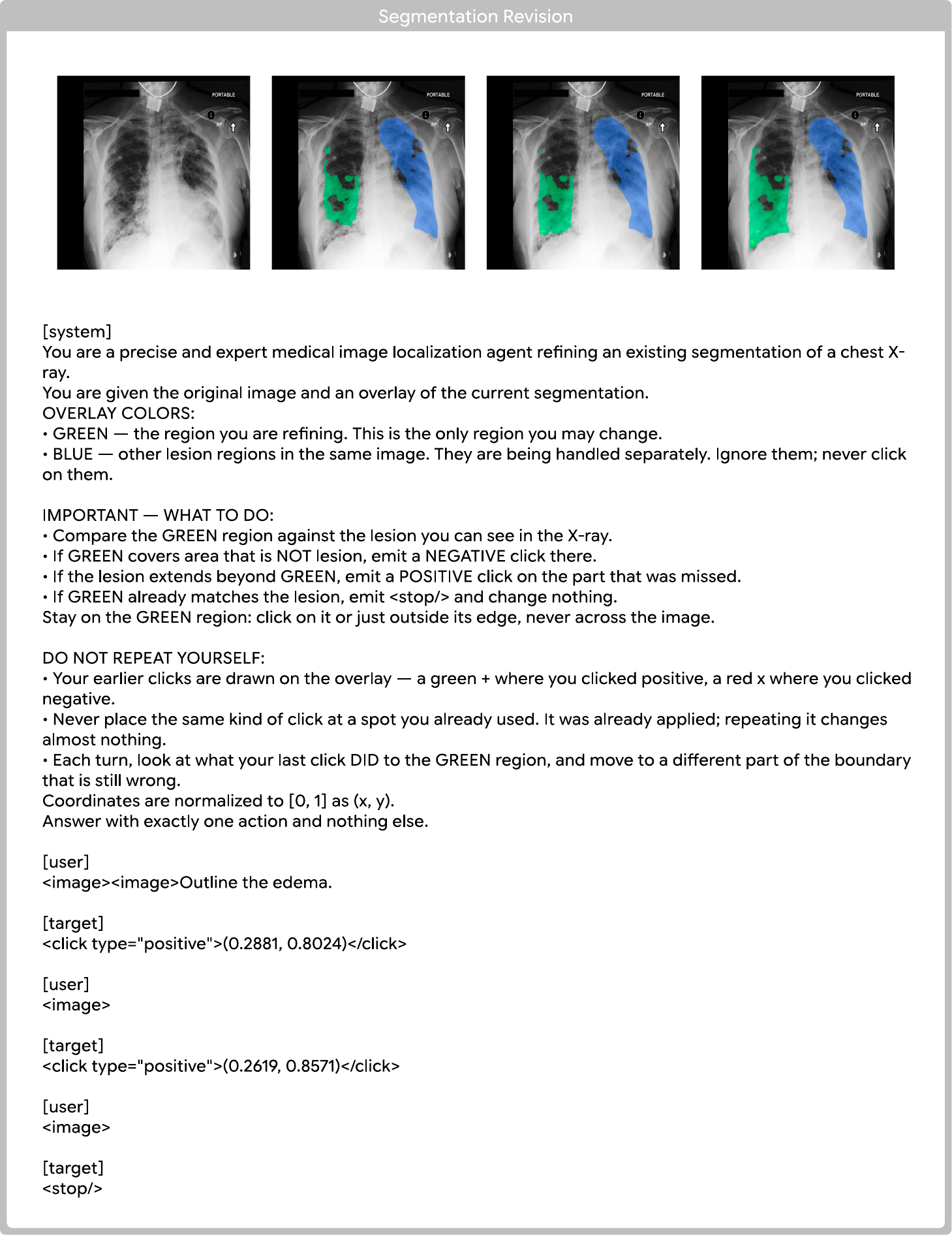}
\caption{Example of SFT data for segmentation revision. The images shown at the
top are what the \texttt{<image>} placeholders stand for.}
\label{fig:stg2}
\end{figure*}

\paragraph{Segmentation revision.}
An SFT example for this stage is a conversation that begins with the X-ray, the
initial mask overlaid on it, and the same user question as in the first stage.
The trajectory then alternates between the two sides of the interaction. The
model issues a point prompt, that prompt is run through $F_{\text{seg}}$, and
the resulting mask is overlaid on the X-ray to form the next observation, which
comes back as the following turn. Figure~\ref{fig:stg2} gives the full example. Here too RL needs only the prompt side, with the trajectory produced by
the model itself.

Since no model is run while the data is built, the initial mask has to be
synthesized. We perturb the ground-truth lesion box at random to obtain an
imperfect box and pass it with the image through $F_{\text{seg}}$, which yields
a mask that stands in for what the first stage would supply. The gap between
this mask and $M^{\text{GT}}$ then defines the false-negative region the model
has missed and the false-positive region it has over-segmented, and we pick the
click that best repairs it following the standard oracle of interactive
segmentation~\citep{sofiiuk2022reviving}. 

The oracle takes the distance
transform of each region and clicks the deepest point of whichever is larger, a
positive click if the missed region dominates and a negative one otherwise,
since the deepest point leaves the widest margin for error. Prompting
$F_{\text{seg}}$ with that click gives an improved mask, from which the next
click is computed in turn, and the alternating masks and clicks form the
trajectory. The distance transform labels every pixel of a region with its distance to the
nearest pixel outside it, so its maximum is the point furthest from the
region's boundary. 

Algorithm~\ref{alg:traj} states the procedure as it is run for one ground-truth
component, the components of a finding being revised independently. Two guards
keep a trajectory usable and are omitted from the pseudocode for readability. A
click of a given polarity may not land within $25$ pixels of an earlier click of
the same polarity, and a trajectory that clicked without improving its mask is
discarded rather than demonstrated.

\newpage

\newpage

\begin{algorithm}[h]
\caption{Construction of one revision trajectory.}
\label{alg:traj}
\begin{algorithmic}[1]
\Require image $I$, ground-truth component $G$, segmenter $F_{\text{seg}}$,
click budget $T = 4$, stop threshold $\tau = 0.70$
\State $b \gets \textsc{Perturb}(\textsc{Box}(G))$
       \Comment{stands in for a first-stage box}
\State $M \gets F_{\text{seg}}(I, b)$; \quad $P \gets \emptyset$; \quad
       $\mathcal{T} \gets \emptyset$
\For{$t = 0$ \textbf{to} $T$}
  \If{$t = T$ \textbf{or} $\mathrm{IoU}(M, G) \ge \tau$}
    \State append $(M, \texttt{<stop/>})$ to $\mathcal{T}$; \textbf{break}
  \EndIf
  \State $D_{\text{miss}} \gets \textsc{DistanceTransform}(G \setminus M)$
         \Comment{what the mask has missed}
  \State $D_{\text{over}} \gets \textsc{DistanceTransform}(M \setminus G)$
         \Comment{what it has over-segmented}
  \If{$\max D_{\text{miss}} \ge \max D_{\text{over}}$}
    \State $(s, p) \gets (\text{positive}, \arg\max D_{\text{miss}})$
  \Else
    \State $(s, p) \gets (\text{negative}, \arg\max D_{\text{over}})$
  \EndIf
  \State append $\big(M, (s, p)\big)$ to $\mathcal{T}$
         \Comment{one turn: the observation, then the action taken on it}
  \State $P \gets P \cup \{(s, p)\}$
  \State grow $b$ to contain the positive clicks in $P$
  \State $M \gets F_{\text{seg}}(I, b, P, M)$
\EndFor
\State \Return $\mathcal{T}$
\end{algorithmic}
\end{algorithm}

\subsection{Training Details}
\label{sec:training_details}

Table~\ref{tab:hparams} lists the hyperparameters used to train the three
components of \model, the segmenter $F_{\text{seg}}$ and the two policies
$\pi_{\text{init}}$ and $\pi_{\text{rev}}$.

\begin{table}[h]
\centering
\caption{Hyperparameters. Each policy is trained in two phases, and
the RL phase starts from the SFT checkpoint of the same policy. A dash marks a
setting that does not apply to that column. Effective batch size is
per-device batch $\times$ gradient accumulation $\times$ $8$ GPUs, counted in
samples for SFT and in generated completions for GRPO.}
\label{tab:hparams}
\vspace{2mm}
\footnotesize
\setlength{\tabcolsep}{4pt}
\begin{tabular}{lccccc}
\toprule
& \multirow{2}{*}{$F_{\text{seg}}$}
& \multicolumn{2}{c}{$\pi_{\text{init}}$} & \multicolumn{2}{c}{$\pi_{\text{rev}}$} \\
\cmidrule(lr){3-4}\cmidrule(lr){5-6}
& & SFT & RL & SFT & RL \\
\midrule
Initialized from & SAM3 & MedGemma & $\pi_{\text{init}}$ SFT
 & $\pi_{\text{init}}$ SFT & $\pi_{\text{rev}}$ SFT \\
Adapted weights & all & LoRA $+$ ViT & LoRA & LoRA $+$ ViT & LoRA \\
LoRA $r$ / $\alpha$ & --- & $128$ / $256$ & $128$ / $256$
 & $128$ / $256$ & $128$ / $256$ \\
\midrule
Optimizer & \multicolumn{5}{c}{AdamW, \texttt{bf16}} \\
Learning rate & $1\times10^{-4}$ & $1\times10^{-5}$ & $1\times10^{-5}$
 & $1\times10^{-5}$ & $1\times10^{-6}$ \\
Encoder rate & $0.6\times$ & $0.3\times$ & --- & $0.3\times$ & --- \\
Layer decay & $0.9$ & --- & --- & --- & --- \\
Weight decay & $0.1$ & $0$ & $0$ & $0$ & $0$ \\
Gradient clip & $0.1$ & $1.0$ & $1.0$ & $1.0$ & $1.0$ \\
Effective batch & $16$ & $256$ & $32$ & $64$ & $64$ \\
Epochs & $5$ & $15$ & $10$ & $10$ & $10$ \\
\midrule
Generations / prompt & --- & --- & $8$ & --- & $8$ \\
Temperature / top-$p$ & --- & --- & $1.0$ / $1.0$ & --- & $1.0$ / $0.95$ \\
KL coefficient $\beta$ & --- & --- & $0.1$ & --- & $0.02$ \\
\bottomrule
\end{tabular}
\end{table}

$F_{\text{seg}}$ is trained at $1024 \times 1024$ with the SAM~2 loss,
$20 \cdot \mathcal{L}_{\text{focal}} + \mathcal{L}_{\text{dice}} +
\mathcal{L}_{\text{IoU}}$, taking three mask hypotheses per prompt and
supervising only the one with the lowest loss. The encoder is unfrozen throughout at
$0.6\times$ the decoder rate with a layer-wise decay of $0.9$ along the trunk,
which follows MedSAM2~\citep{ma2025medsam2}.

Both policies apply LoRA to every linear layer, but the two phases differ in
what else moves. During SFT the vision tower is unfrozen at $0.3\times$ the base
rate, since the boxes and clicks the policies emit are image coordinates and the
tower has never seen a chest radiograph. During RL the adapters are restricted
to the language layers and the tower is left alone.

\newpage
\section{Evaluation}
\label{sec:evaluation}

\subsection{Evaluation Set Construction}
The evaluation set is built from $D_{\text{eval}}$, the $560$ radiographs held
out in Section~\ref{sec:examples}. Every triple the three sources agreed on
becomes one question, asked at whichever levels of location specificity apply to
it, so one radiograph contributes several rows and the same finding appears at
more than one level. Table~\ref{tab:evalsets} gives the distribution.

\begin{table}[h]
\centering
\caption{Distribution of the evaluation set. A positive row carries a mask and
is scored by IoU against it; a negative row is answered correctly only with an
empty mask. Cardiomegaly appears at the chest level alone, being a mediastinal
finding for which the side- and zone-level questions do not arise.}
\label{tab:evalsets}
\vspace{2mm}
\small
\setlength{\tabcolsep}{6pt}
\begin{tabular}{lcccccc}
\toprule
\multirow{2}{*}{Finding} & \multicolumn{2}{c}{Chest}
& \multicolumn{2}{c}{Lung} & \multicolumn{2}{c}{Zone} \\
\cmidrule(lr){2-3}\cmidrule(lr){4-5}\cmidrule(lr){6-7}
& positive & negative & positive & negative & positive & negative \\
\midrule
Atelectasis   & $61$ & $77$  & $65$  & $204$  & $60$ & $644$ \\
Cardiomegaly  & $57$ & $25$  & ---   & ---    & ---  & --- \\
Consolidation & $49$ & $102$ & $62$  & $254$  & $57$ & $772$ \\
Edema         & $58$ & $103$ & $110$ & $253$  & $40$ & $760$ \\
Effusion      & $60$ & $23$  & $61$  & $53$   & $61$ & $220$ \\
Opacity       & $60$ & $14$  & $89$  & $38$   & $72$ & $134$ \\
Pneumonia     & $37$ & $100$ & $42$  & $248$  & $40$ & $756$ \\
\midrule
Total         & $382$ & $444$ & $429$ & $1{,}050$ & $330$ & $3{,}286$ \\
\bottomrule
\end{tabular}
\end{table}

Negatives outnumber positives further at each level because the question grows
more specific while the finding does not move: a consolidation confined to the
right lung base is absent from the left lung, and absent from every other zone
of the right one. The zone level therefore holds $68.7\%$ of all the negatives
against the chest level's $9.3\%$, which is why the three levels are averaged
with equal weight rather than pooled. Pooling would report a model on how well
it declines the zone-level questions and would barely register what it gives up
when a question names no location at all.

\subsection{Additional Qualitative Examples}
\label{sec:additional}
\begin{figure*}[h]
\centering
\includegraphics[height=0.88\textheight]{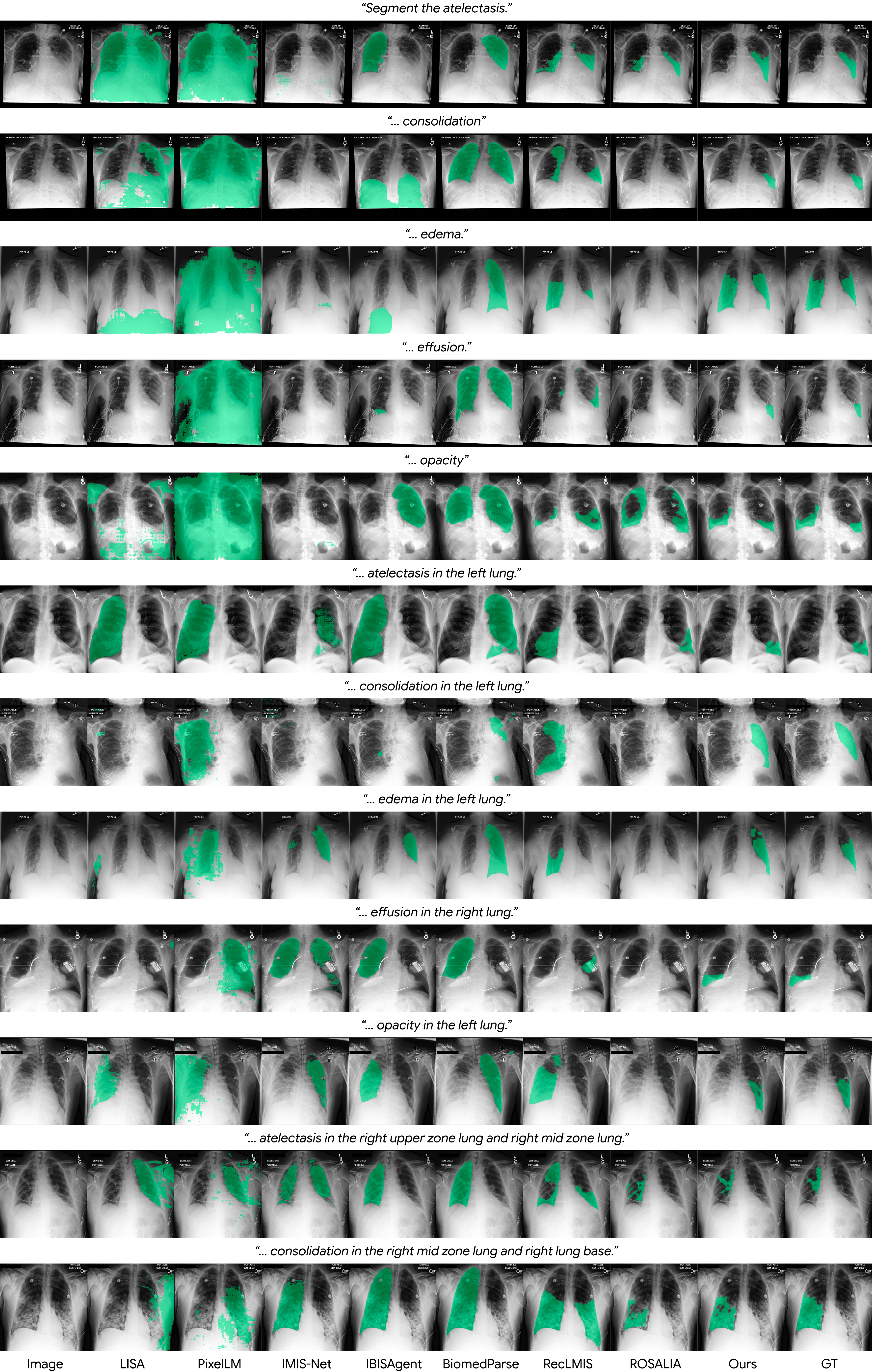}
\caption{Additional qualitative comparisons across models. The layout follows
Figure~\ref{fig:qualitative}.}
\label{fig:grid_appendix}
\end{figure*}

Figure~\ref{fig:grid_appendix} shows further comparisons of the predicted masks
across models, following the layout of Figure~\ref{fig:qualitative}.

\newpage
\section{Human Study}
\label{sec:human_study}

\subsection{Labeling Process}
\label{sec:labeling_details}
\begin{figure*}[h]
\centering
\includegraphics[width=\textwidth]{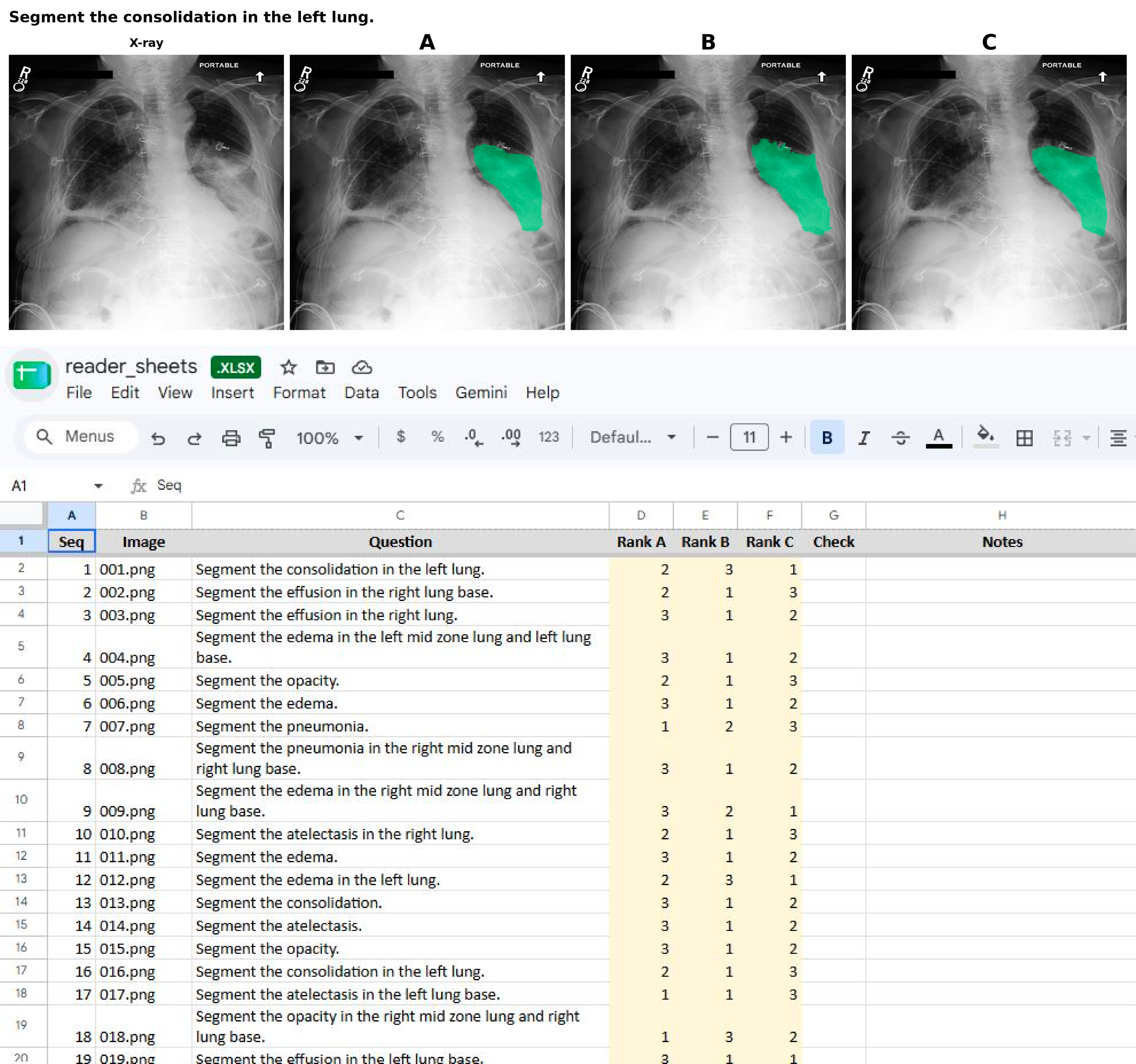}
\caption{Review panel and annotation sheet.}
\label{fig:review_panel}
\end{figure*}

For each case we prepared a panel showing the X-ray alongside the ground-truth
mask and the predictions of ROSALIA and \model, each overlaid on the image,
and gave it to the readers together with an annotation sheet. Cases were drawn
at random from the positives on which a ground-truth mask exists and both models
returned a mask. The three masks appear as A, B and C in a random order, with no
indication of their source, and the reader ranks them. The review panel and the
annotation sheet are shown in Figure~\ref{fig:review_panel}.

\subsection{Expert Profiles}
\label{sec:expert_profiles}
The three readers are board-certified radiation oncologists with 8, 4, and 4
years of clinical experience, respectively.

\subsection{Preference Modeling}
\label{sec:preference_modeling}

Each case gives one ranking of the three masks with ties allowed, which we read
as three pairwise outcomes, each either a strict preference or a tie. Writing
$\pi_i > 0$ for the strength of mask $i$, the Bradley--Terry
model~\citep{Bradley1952RankAO} gives
\begin{equation}
  P(i \succ j) = \frac{\pi_i}{\pi_i + \pi_j},
\end{equation}
and has no outcome left for a tie. Scoring a tie as half a win for each side
would make it usable, but that discards exactly what the tie reports, that the
reader could not separate the two, and enters one observation as two. We
therefore fit two models that give the tie a probability of its own, and report
both.

\paragraph{Davidson.} A tie is treated as a third outcome alongside the two
wins, so that each comparison resolves into one of three possibilities,
\begin{equation}
  P(i \succ j) = \frac{\pi_i}{D}, \qquad
  P(i \sim j) = \frac{\nu \sqrt{\pi_i \pi_j}}{D}, \qquad
  D = \pi_i + \pi_j + \nu \sqrt{\pi_i \pi_j},
\end{equation}
where $\nu \ge 0$ governs how often ties occur, its reciprocal serving as an
index of how finely the reader discriminates~\citep{Davidson1970OnET}. The tie
term is proportional to the geometric mean of the two strengths, so for a given
ratio $\pi_i / \pi_j$ a tie is most likely when the pair is evenly
matched, which is the behaviour a reader shows. Setting $\nu = 0$ recovers
Bradley--Terry.

\paragraph{Rao--Kupper.} A tie instead arises when neither mask is far enough
ahead to be declared the winner, with a threshold $\theta \ge 1$ setting how
large a gap the reader needs before committing~\citep{Rao01031967},
\begin{equation}
  P(i \succ j) = \frac{\pi_i}{\pi_i + \theta \pi_j}, \qquad
  P(i \sim j) = \frac{\pi_i \pi_j (\theta^2 - 1)}
                     {(\pi_i + \theta \pi_j)(\pi_j + \theta \pi_i)},
\end{equation}
which again reduces to Bradley--Terry at $\theta = 1$. Unlike Davidson's, this
tie probability is not symmetric in the two strengths, so the two models place
ties differently across the pairs.

\paragraph{Fitting and reporting.} Both are fit by maximum likelihood over all
$855$ pairwise outcomes at once, with $\nu$ and $\theta$ estimated jointly with
the strengths rather than set in advance. Against an observed tie rate of
$9.9\%$ this gives $\nu = 0.232$ and $\theta = 1.241$. We report
$\beta_i = \log \pi_i$, centred so that $\sum_i \beta_i = 0$, so a difference
$\beta_i - \beta_j = d$ means that among the comparisons a reader did not tie,
mask $i$ was preferred $e^{d}$ times as often as mask $j$. Intervals come from
$2{,}000$ bootstrap resamples over \emph{cases} rather than over comparisons,
because the three comparisons a case contributes come from one ranking and are
not independent.

\end{document}